\documentclass[conference]{IEEEtran}
\IEEEoverridecommandlockouts

\usepackage{cite}
\usepackage{latexsym}

\usepackage{amssymb}
\usepackage{amsmath}
\usepackage{amsthm}
\usepackage{booktabs}
\usepackage{enumitem}
\usepackage{graphicx}
\usepackage{color}
\usepackage{multirow}
\usepackage{pifont}
\usepackage{bm}
\usepackage{newunicodechar}
\newunicodechar{✓}{\ding{51}}
\newunicodechar{✗}{\ding{55}}
\usepackage{xcolor}
\usepackage{hyperref}
\usepackage{tabularx}
\usepackage{algorithm}
\usepackage{booktabs}
\usepackage{mathrsfs}
\usepackage{siunitx}
\usepackage{algcompatible}
\usepackage{listings}
\usepackage{float}
\usepackage{placeins}
\usepackage{textcomp}
\usepackage{balance}

\definecolor{darkgreen}{rgb}{0.0, 0.5, 0.0}
\algnewcommand{\CommentX}[1]{\Statex \textcolor{darkgreen}{$\triangledown$ \emph{#1}}}
\algnewcommand\algorithmicreturn{\textbf{return}}
\algnewcommand\RETURN{\State \algorithmicreturn}

\def\BibTeX{{\rm B\kern-.05em{\sc i\kern-.025em b}\kern-.08em
    T\kern-.1667em\lower.7ex\hbox{E}\kern-.125emX}}

\begin{document}

\title{DCoPilot: Generative AI-Empowered Policy Adaptation for Dynamic Data Center Operations}

\author{\IEEEauthorblockN{Minghao Li, Ruihang Wang*\thanks{*Corresponding author. This work was conducted while Ruihang Wang was a research fellow at NTU. He is now with Yunnan University.}, Rui Tan, Yonggang Wen}
\IEEEauthorblockA{Nanyang Technological University, 
Singapore\\
E-Mail: \{minghao002, ruihang001\}@e.ntu.edu.sg, \{tanrui, ygwen\}@ntu.edu.sg}}

\maketitle

\begin{abstract}
Modern data centers (DCs) hosting artificial intelligence (AI)-dedicated devices operate at high power densities with rapidly varying workloads, making minute-level adaptation essential for safe and energy-efficient operation.
However, manually designing piecewise deep reinforcement learning (DRL) agents cannot keep pace with frequent dynamics shifts and service-level agreement (SLA) changes in an evolving DC.
This specification-to-policy latency causes a lack of timely, effective control policies, which may lead to service outages.
To bridge the gap, we present DCoPilot, a hybrid framework for generative control policies in dynamic DC operation. 
DCoPilot synergizes two distinct generative paradigms, i.e., a large language model (LLM) that performs symbolic generation of structured reward forms, and a hypernetwork that conducts parametric generation of policy weights.
DCoPilot operates through three coordinated phases: (i) simulation scale-up, which stress-tests reward candidates across diverse simulation-ready (SimReady) scenes; (ii) meta policy distillation, where a hypernetwork is trained to output policy weights conditioned on SLA and scene embeddings; and (iii) online adaptation, enabling zero-shot policy generation for updated specifications via interpolation within the trained specification envelope.
Evaluated across five control task families spanning diverse DC components, DCoPilot achieves near-zero constraint violations and outperforms all baselines across specification variations. Ablation studies validate the effectiveness of LLM-based unified reward generation in enabling stable hypernetwork convergence.
\end{abstract}


\section{Introduction}
The proliferation of artificial intelligence (AI) applications is spurring global computing needs, where capital investment in data centers (DCs) will expand at 21\% annually through 2029~\cite{DellOro2025Capex}.
This growing investment incentivizes the development of AI-dedicated data centers (AIDCs). These facilities are characterized by high-density racks with high-performance GPU~\cite{Uptime2025AISurvey}.
Unlike standard computational tasks, hosting AI training tasks with advanced graphics processing unit (GPU) servers is projected to raise average rack power density from \SI{8}{\kilo\watt} to \SI{30}{\kilo\watt} by 2027~\cite{McKinsey2024AIDCGrowth}. 
At such densities, iteration-scale GPU workload behavior produces large idle-to-peak power swings that translate into tremendous thermal fluctuations~\cite{Google2025Fluctuations}.

AIDC's high power density and bursty compute loads increase energy consumption and reduce cooling response time for reliable operations.
In practice, operators are required to manage these dynamics on minute-scale control horizons~\cite{DeepMind2018Cooling}.
Meanwhile, frequent server installations and changing tenant service-level agreements (SLAs) within a shared data hall further affect the steadiness of coordination between information technology equipment (ITE) and cooling facilities~\cite{wustenhoff2002service}.
However, manually engineering setpoints and rule-based playbooks is time-consuming and cannot keep pace with the tight response window, as it requires continual retuning across thermal–electrical variables~\cite{lazic2018data}.
As a cautionary example, a planned cooling system upgrade at an Equinix Singapore DC in October 2023 manually misconfigured the chilled-water system, causing overheating and bank service disruption~\cite{Moss2023EquinixSingapore}.
Consequently, AIDC operations require policies that can adapt quickly and safely to evolving DC scenarios and varied tenant-specific SLAs.

\begin{figure}[t]
    \centering
    \includegraphics[width=\linewidth]{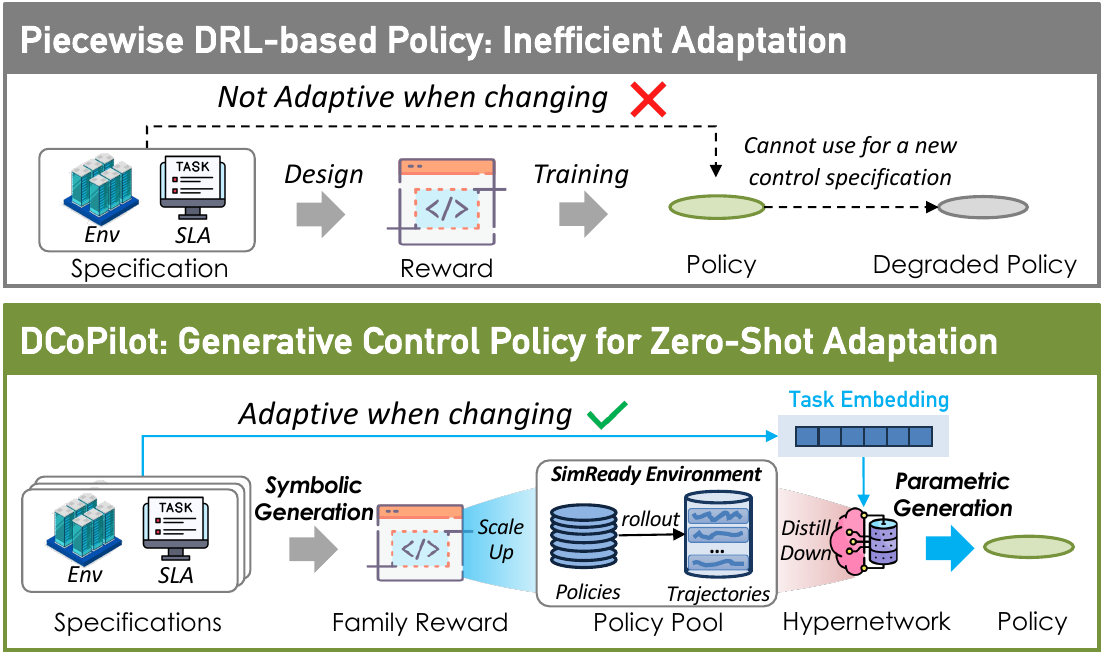}
    \vspace{-2em}
    \caption{DCoPilot replaces manual, piecewise DRL-based policies with a generative pipeline: an LLM formalizes operational \textit{specifications} into a symbolic family reward, and a hypernetwork generates parametric \textit{policy} network weights under evolving environments.}
    \label{fig:intro}
\end{figure}

Existing control policies for cyber-physical systems (CPS) can be categorized into non-generative approaches, which are designed for static environments, and generative policies that adapt to varying specifications. 
Among non-generative methods, expert-engineered policies rely on human expertise and experience to manually tune parameters for a specific control scenario. 
In DC operations, a proportional-integral-derivative (PID) controller is employed to adjust the server's internal fan speed in response to its temperature, where engineers need to manually tune the parameters to achieve a stable and efficient response~\cite{durand2013data}.
Data-driven controllers under fixed objectives learn policies against pre-specified costs or constraints, but still depend on domain-specific modeling and reward shaping.
Model-based approaches identify a dynamics surrogate and solve a constrained optimization via model predictive control (MPC) to trade off energy against temperature bounds~\cite{beitelmal2006model,lazic2018data}.
To reduce the modeling efforts, model-free approaches directly optimize expected return by interacting with the environment to train deep reinforcement learning (DRL) agents~\cite{moriyama2018reinforcement,li2019deep, ong2021deep}.
Recent work fuses physics priors with cross-environment adaptation, where an offline thermodynamics-aware model pairs with safe online fine-tuning for DRL agents~\cite{wang2023phyllis}.
Despite these advances, manual modeling and reward shaping remain expert-intensive, creating specification-to-policy latency that delays control policy deployment and may precipitate service disruptions.

In contrast, generative policies aim to overcome these limitations by synthesizing control strategies conditioned on goals, constraints, and contextual inputs. 
Early studies adopt the large language models (LLMs) as direct policy generators, relying on in-context expert knowledge for industrial control \cite{song2023pre}.
However, LLM hallucinations may recommend risky actions, leading to computing service outages.
To ensure the safety, some LLM-based control policy generation approaches focus on industrial control code generation, involving retrieval augmented generation (RAG) with knowledge base to generate control codes based on user prompts \cite{koziolek2024llm, fakih2024llm4plc}. 
However, LLM-generated PID control is limited to setpoint tracking and cannot optimize the objectives of DC operations.
Recent studies focus on using LLMs for DRL agent training.
These approaches mainly focus on robotics problems using evolutionary gradient-free approaches to design reward functions \cite{song2023self, ma2023eureka}.
Unlike static embodied robots, DCs evolve due to ITE upgrades and SLA changes by tenants.
The fixed parameters of LLM-guided DRL policies cannot generalize to skewed distributions in dynamic DC operations.
In sum, existing policies are tailored to fixed operational specifications without considering the rapidly changing DC environments and tenant SLAs.

To bridge the gap, we introduce DCoPilot, a hybrid framework for generative control policies in dynamic DC operation. 
As shown in Figure~\ref{fig:intro}, the core contribution of our work lies in the synergistic integration of two forms of generative AI (i.e., LLM and hypernetwork~\cite{ha2016hypernetworks}) to enable zero-shot policy generation through interpolation over offline-trained policies under evolving DC specifications, including shifting system configurations and SLA setpoints.
The LLM defines what to do, and the hypernetwork instantly creates a policy for achieving it.
In particular, an LLM generates a shared reward form for a family of Markov decision processes (MDPs), compiling operator prompts, domain-specific metrics, and SLA bands.
A hypernetwork translates the task embedding of an operational specification into deployment-ready policy weights, addressing the LLM's limitations with continuous numerical parameters. 

A key design principle of DCoPilot is to shift all training overhead offline through LLM-based automation, while the online phase leverages the hypernetwork for instant policy generation, addressing the challenge of specification-to-policy latency.
At its core, DCoPilot generates DC control policy in three steps: (i) simulation scale-up, (ii) meta policy distillation, and (iii) online adaptation. 
During the offline simulation scale-up, DCoPilot leverages a Reward LLM to generate reward candidates and reflect on trajectories of generated candidates training over the min/max boundary operational specifications (i.e., DC system and tenant SLA) in the synthetic simulation-ready (SimReady)~\cite{nvidia_simready_assets} environment.
Then, DCoPilot selects the best reward form candidates, trains multiple DRL agents for varied operational specifications, and rollouts a near-optimal trajectory pool.
In the meta policy distillation stage, a hypernetwork is trained on the trajectory pool, encoding each specification as a unique task embedding to capture shared policy structures.
During online adaptation, new operational specifications are used as input to the pre-trained hypernetwork to generate policy weights for real-time control adjustment.

We evaluate DCoPilot against three LLM-based control baselines and one ablation setting across five task families spanning diverse DC components and operational objectives.
DCoPilot achieves the lowest violation cost below \SI{0.2}{\celsius} across all task families, significantly outperforming baselines that range from \SI{0.77}{\celsius} to \SI{8.78}{\celsius}, and maintains consistent temperature control with mean errors below \SI{0.5}{\celsius} across different SLA bounds and server configurations.
In a 40-day operational test with two specification changes, DCoPilot keeps violations near zero throughout, while task-specific DRL methods experience large violation spikes up to \SI{4}{\celsius} during their multi-day retraining periods.
Compared with meta-learning approaches on zero-shot adaptation, DCoPilot exhibits more stable control across power and temperature metrics during specification transitions, without needing any online fine-tuning.
Ablation studies further show that DCoPilot's LLM-based generation of unified family reward form makes hypernetwork training stable.

Our contributions are summarized as follows:
\begin{itemize}
    \item We propose DCoPilot, the first framework integrating symbolic reward and parametric policy network generation for dynamic DC operations, establishing a new paradigm that combines LLMs with hypernetworks for adaptive CPSs.
    \item We design a dual-stage generative pipeline that synergizes LLM-based reward formulation with hypernetwork-based policy generation, confining training to offline phases while enabling zero-shot online adaptation.
    \item Through extensive experiments across five control task families, we demonstrate superior efficiency and policy quality over baselines, with an ablation study validating a significant reduction in specification-to-policy latency.
\end{itemize}

{\em Paper organization:} \textsection\ref{sec:related} reviews related work. \textsection\ref{sec:pre} presents preliminaries and problem definition. \textsection\ref{sec:design} presents the design of DCoPilot. \textsection\ref{sec:eval} presents evaluation results. \textsection\ref{sec:discuss} discusses several related issues. \textsection\ref{sec:conclude} concludes this paper.

\section{Related Work}
\label{sec:related}
\subsection{DC Operation Optimization}
DC operation optimization aims to enhance energy efficiency, reduce costs, and improve sustainability in large-scale computing facilities. 
Traditional approaches employ model-based techniques such as model predictive control for regulating cooling systems~\cite{lazic2018data} and evolutionary algorithms for geographical workload optimization~\cite{pasricha2023mosaic}.
More recent approaches apply DRL to handle dynamic DC conditions. 
DeepEE~\cite{ran2019deepee} proposes a deep Q-network framework that jointly optimizes job scheduling and cooling control in a DC simulation.
Wang et al. demonstrate that DRL-based joint IT-facility optimization achieves energy reductions \cite{wang2021joint}.
Sarkar et al.~\cite{sarkar2024sustainability} propose a multi-agent RL digital twin environment for DCs enabling joint optimization of IT servers, cooling, load-shifting and battery storage to reduce carbon footprint.
More recently, Zhan et al. employ offline RL with graph neural networks (GNN) to achieve 14-21\% energy reductions by modeling physical dependencies \cite{zhan2025cooling}.
However, these methods often require extensive domain expertise, suffer from high training costs, and lack robustness in rapidly evolving scenarios with varying SLAs and installed servers. As DCs scale to uphold AI applications that process massive data from the edge, current techniques fall short in providing automated, adaptive, and cost-effective solutions for future requirements.

\subsection{Few-Shot Adaptation for CPS Control}

\begin{figure}[t]
    \centering
    \includegraphics[width=\linewidth]{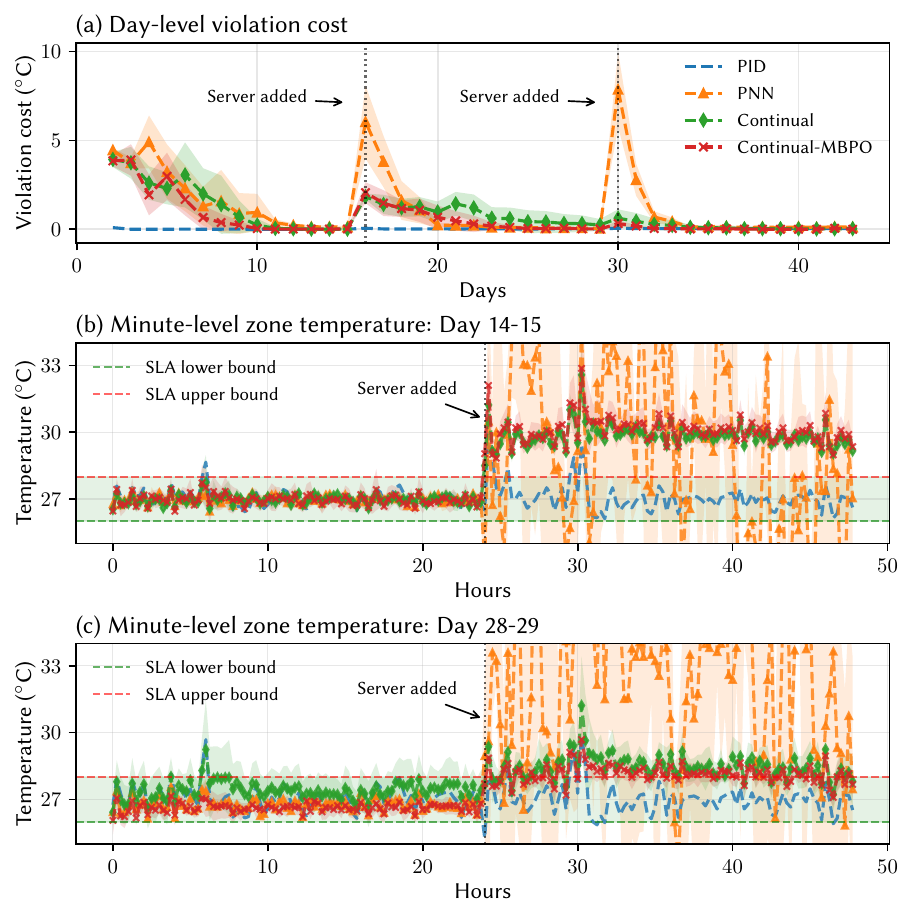}
    \vspace{-2.5em}
    \caption[Adaptation performance of continuous learning approaches at day-level and minute-level evaluations.]
    {Adaptation performance of continuous learning approaches at day-level and minute-level evaluations.  Day-level violation cost is the time-average magnitude by which the zone temperature exceeds the SLA band, i.e., $\bar v_{\mathrm{day}}=\frac{1}{N_{\mathrm{day}}}\sum_{t\in\mathrm{day}} \max\{0,\,T_t-T_\text{U},\,T_\text{L}-T_t\}$, where $T_\text{U}$ and $T_\text{L}$ are the upper and lower SLA bounds, and $N_{\mathrm{day}}$ is the number of steps. }
    \label{fig:day_level_minute_level}
\end{figure}


Adaptive CPS control includes meta-learning for few-shot adaptation, continual learning for sequential task acquisition, and model-based reinforcement learning for sample-efficient policy optimization. Model-agnostic meta-learning (MAML) learns initialization parameters enabling rapid adaptation through gradient steps, while probabilistic embeddings for actor-critic learning (PEARL) infers task embeddings from trajectories to condition policies~\cite{finn2017model,rakelly2019efficient}. Progressive neural networks (PNN) address continual learning by adding network columns for each task with lateral connections to prevent catastrophic forgetting~\cite{rusu2016progressive}, while model-based policy optimization (MBPO) achieves sample efficiency by learning dynamics models to generate synthetic training data~\cite{janner2019trust}. These methods have shown promise across CPS-related domains, where Meta-Controller achieves few-shot imitation learning~\cite{cho2024meta}, LEGION enables skill recombination for lifelong learning~\cite{meng2025preserving}, and Phyllis incorporates physical priors for efficient adaptation in DCs~\cite{wang2023phyllis}. 
While meta learning requires few-shot examples from new tasks to adapt, generative AI operates in a zero-shot manner, leveraging its pre-trained knowledge to generate actions without waiting for environmental feedback. 
Figure~\ref{fig:day_level_minute_level} demonstrates these cases of using continual learning approaches for DC operation.
In DC operations governed by minute-level decision windows, the online exploration and data collection required by few-shot adaptation may lead to computing performance degradation or even outage.
In sum, few-shot adaptation approaches only take fixed objective functions and cannot adapt instantaneously to burst workloads.

\subsection{Generative AI for Adaptive Control}
Generative AI encompasses diverse architectures, capable of directly synthesizing novel outputs from learned distributions. 
Recent advances have explored three paradigms for policy generation, including diffusion models for action sampling, hypernetworks for parametric policy adaptation, and LLMs for symbolic reasoning. 

Diffusion models formulate policy learning as an iterative denoising process, enabling multi-modal action generation in high-dimensional spaces~\cite{ho2020denoising}.
While Diffusion Policy~\cite{chi2025diffusion} achieves strong results in robotic manipulation with multiple valid actions per state, DC control is different.
In DC control, per-step actions are constrained by SLAs and actuator limits, and performance depends on constraint-satisfying behavior rather than sampling diverse instantaneous actions.

Hypernetworks generate weights for target networks conditioned on task embeddings, enabling rapid adaptation across task distributions~\cite{ha2016hypernetworks}. 
In multi-task reinforcement learning, hypernetworks have been applied to generate task-specific policy parameters, demonstrating effective knowledge transfer and few-shot adaptation~\cite{beck2023survey}. 
Recent work extends hypernetworks to continuous control, where conditioning on goal embeddings enables zero-shot generalization to unseen objectives~\cite{rezaei2023hypernetworks}. 
However, existing hypernetwork approaches lack the symbolic reasoning to interpret evolving operational requirements, making them hard to align control policies with operational intents.

LLMs have demonstrated capabilities in code generation and logical reasoning, making them suitable for generating control policies or reward functions. Early studies leverage LLMs as direct policy generators through in-context learning for industrial control~\cite{song2023pre}. However, LLM hallucinations pose severe operational risks in safety-critical systems. To constrain the LLM's output and reduce hallucinations, recent works use knowledge bases to guide the generation of industrial control code~\cite{koziolek2024llm, fakih2024llm4plc}. 
However, these studies focus on a fixed optimization objective.
Another direction employs LLMs for reward function design in DRL training. Methods like Eureka~\cite{ma2023eureka} and its variants~\cite{ma2024dreureka} use evolutionary gradient-free approaches to optimize reward code for robotics tasks. 
However, these methods require extensive training for each new specification, creating specification-to-policy latency way beyond the minute-scale DC control horizons.

In sum, existing generative approaches address only isolated aspects of the control problem and cannot simultaneously interpret operational intent and synthesize adaptive policies.  
This motivates a synergized generative framework where the LLM defines what to do, and the hypernetwork creates a policy to achieve it.

\section{Preliminaries \& Problem Definition}
\label{sec:pre}
In this section, we introduce the preliminaries of air-cooled DC operation and formally define the problem of generative policy synthesis for dynamic DC control.

\subsection{DC Architecture \& Operation}

\begin{figure}[t]
    \centering
    \includegraphics[width=\linewidth]{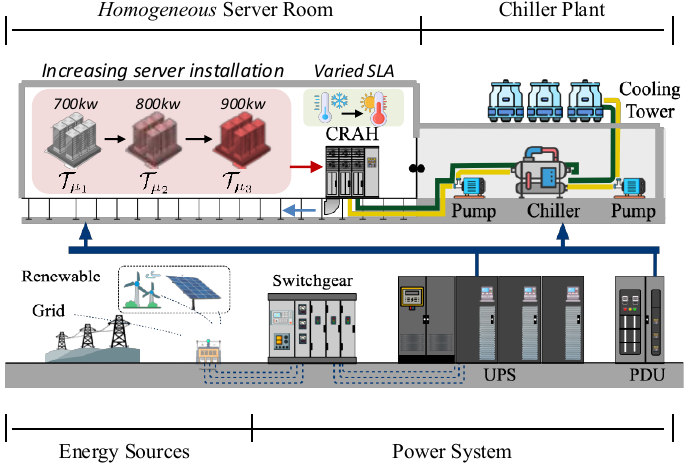}
    \vspace{-2em}
    \caption{General topology of DC infrastructure, including server rooms, chiller plant, electrical system, and power grid. DC operators continuously install new servers and adjust SLAs, making \textit{homogeneous} changes over time.}
    \label{fig:dc_stages}
\end{figure}

\subsubsection{DC system architecture}
We consider a typical enterprise DC equipped with an air-cooled system, the most widely deployed cooling architecture. The infrastructure comprises server rooms with IT equipment (ITE), computer room air conditioning (CRAC) units, chillers, cooling towers, and the electrical distribution system.
The cooling process consists of two stages. 
The first stage adopts the CRACs to supply cold air to the IT devices and cool the return hot air.
The second stage transfers the water-carried heat by the chiller and dissipates it to the ambient.
This two-stage architecture enables independent control of indoor air conditions and outdoor heat rejection. The control system manages continuous SLA setpoints to maintain ITE performance while achieving sustainability goals.

DC operators continuously install new servers to meet growing computational demands, as illustrated in Figure~\ref{fig:dc_stages}. 
This server expansion represents a homogeneous change, where a transformation preserves the system structure while scaling its parameterized heat gains. Specifically, adding servers to existing racks maintains the same cooling infrastructure topology, i.e., identical numbers of CRACs, chillers, and cooling towers, preserves the same state space dimensions, and keeps the same action space dimensions.  
For instance, upgrading from 10 to 15 servers per rack increases total heat load from \SI{700}{\kilo\watt} to \SI{900}{\kilo\watt}, requiring a new optimized control policy, yet the action space and underlying physical relationships remain structurally identical. 

\begin{figure}[t]
    \centering
    \includegraphics[width=\linewidth]{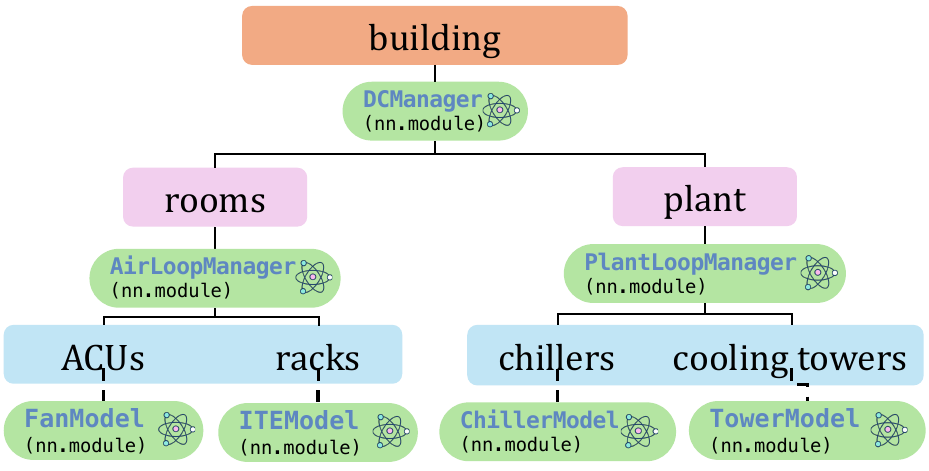}
    \vspace{-2em}
    \caption{The hierarchical structure of a DC SimReady scene. A DC SimReady scene consists of three-level SimReady assets. Each SimReady asset is a surrogate model trained from operational data.}
    \label{fig:dtfile}
\end{figure}

\subsubsection{Hierarchical SimReady scene}
We adopt a hierarchical system model for a DC that is purely data-driven, developed in our prior work~\cite{wang2026toward}.
The model is a SimReady digital scene~\cite{nvidia_simready_assets} that predicts future states from current states and actions.
As illustrated in Figure~\ref{fig:dtfile}, a scene description file specifies the configuration of all assets and their interconnections, resulting in a multi-level structure. 
The top level represents the overall \texttt{Building}. The second level separates \texttt{Rooms} from \texttt{Plant}. For the third level, we focus on \texttt{Racks} and \texttt{ACUs} within \texttt{Rooms}, and \texttt{Chillers} and \texttt{Towers} within \texttt{Plant}. An air-cooling unit (ACU) conditions and delivers supply air for ITE. A chiller removes heat from a secondary water loop to provide chilled water to ACUs. A cooling tower rejects condenser heat via evaporative exchange. 

Each component in a DC can be modeled as an asset paired with a learned surrogate trained from operational data. 
Coupling among assets is handled by its corresponding manager classes, where \texttt{AirLoopManager} exchanges air-side variables for server rooms, \texttt{PlantLoopManager} coordinates water-side flows for the chiller plant, and the top-level \texttt{DCManager} supervises the global simulation of the entire DC. 
With these SimReady assets composed through this hierarchy, the resulting SimReady scene provides a foundation for simulation and serves as a basis for generating synthetic scenes.
These learned model-free surrogates preclude direct application of MPC-style controllers.

\subsubsection{DRL-based cooling control}
The facility controls perform with a continuous action space.
For instance, the DC cooling control aims to improve power efficiency while maintaining operating temperature at a target setpoint.
This operation periodically adjusts the air temperatures and mass flow rates supplied from the CRAC units, which can be modeled as a constrained MDP and solved with DRL techniques \cite{li2019transforming, ran2019deepee, van2019control}.
Formally, the DC cooling control problem is represented as an MDP:
\begin{equation}
\mathcal{M} = (\mathcal{S}, \mathcal{A}, \mathcal{T}, \mathcal{R}),
\end{equation}
where $\mathcal{S}$ denotes the state space capturing thermal and facility conditions, 
$\mathcal{A}$ is the action space consisting of continuous control inputs for active cooling facilities, 
$\mathcal{T}:\mathcal{S}\times \mathcal{A} \rightarrow \text{Dist}(\mathcal{S})$ models the stochastic transition dynamics, 
and $\mathcal{R}: \mathcal{S}\times \mathcal{A} \rightarrow \mathbb{R}$ defines the reward function that penalizes SLA violations and incentivizes sustainability goals.

\begin{figure*}[t]
    \centering
    \includegraphics[width=0.98\linewidth]{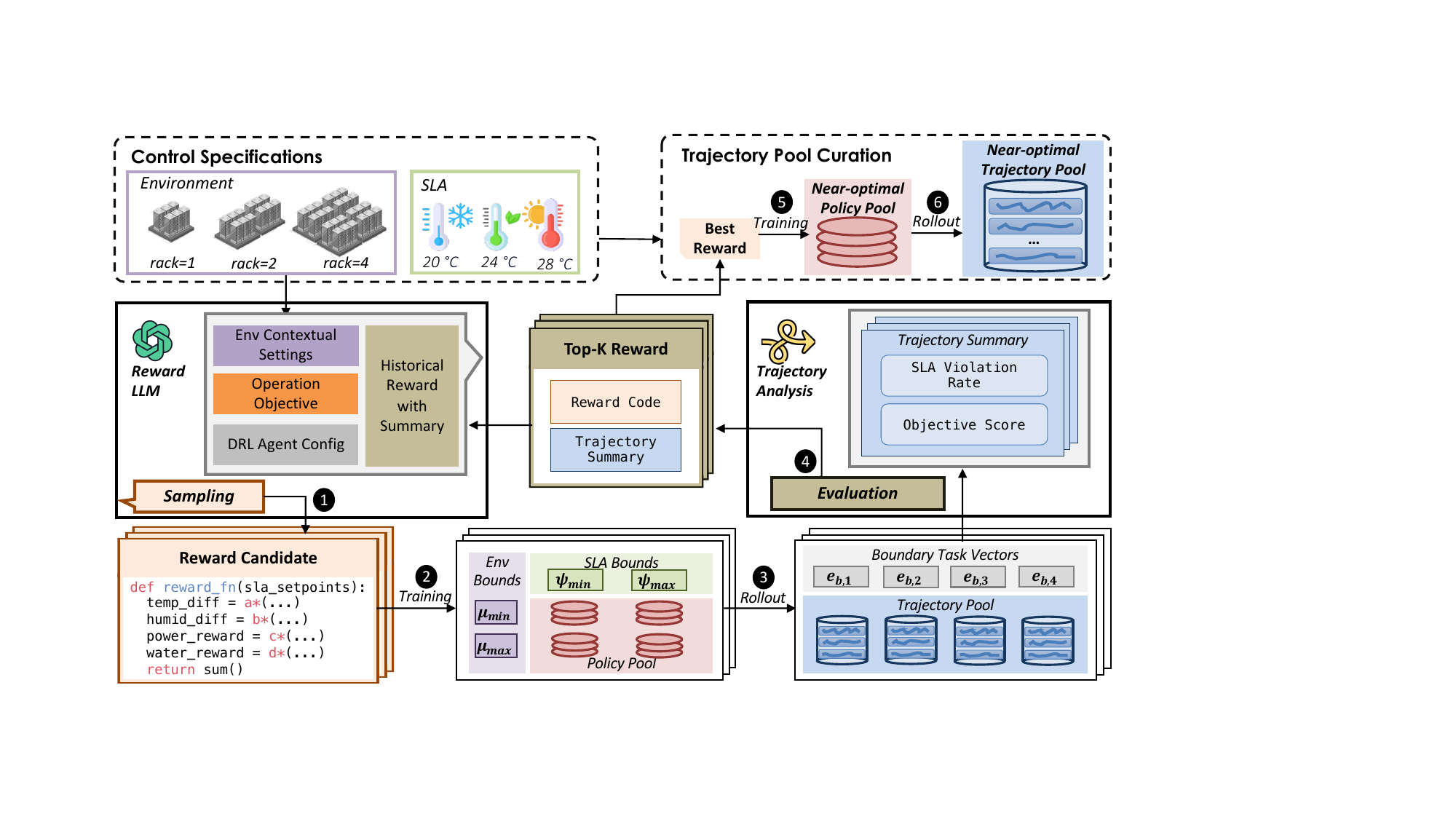}
    \vspace{-1em}
    \caption{ 
    Given scene $G$ and operator objective $I_{\text{ops}}$, the Reward LLM generates reward code candidates (\ding{202}). Each candidate is evaluated on boundary conditions by training policies and collecting trajectories (\ding{203}-\ding{204}). Then, it ranks candidates based on SLA violation costs and objective scores, selecting top-$k$ for evolutionary refinement (\ding{205}). The best reward form $\bar{R}^*$ trains DRL agents across sampled specifications $\mathbf{e} = \bm{\mu} \otimes \bm{\psi}$ (\ding{206}), generating near-optimal trajectory pool $\mathbf{T}$ for hypernetwork distillation (\ding{207}).}
    \label{fig:workflow}
\end{figure*}

\subsection{Problem Formulation}
\subsubsection{Specification-driven MDP family}
While DC operations involve diverse variations, we focus on operational changes that can be naturally parameterized. These variations change the MDP with every new operational specification.
Installing servers modifies thermal dynamics ($\mathcal{T} \to \mathcal{T}_{\mu}$), parameterized by $\mu$ capturing power density. SLA updates modify reward functions ($\mathcal{R} \to \mathcal{R}_{\psi}$), parameterized by $\psi$ encoding SLA setpoints (e.g., temperature and humidity). This defines a family of parameterized MDPs:
\begin{equation}
\mathscr{M} = \{\mathcal{M}_i | \mathcal{M}_i = (\mathcal{S}, \mathcal{A}, \mathcal{T}_{\mu_i}, \mathcal{R}_{\psi_i})\},
\end{equation}
where $\mu_i \sim p(\mu)$ and $\psi_i \sim p(\psi)$. State and action spaces remain fixed across the family, reflecting homogeneous changes.

\subsubsection{The dual challenge}
Achieving generative policies for dynamic DCs faces two intertwined challenges:

\textbf{Reward Form Heterogeneity (C1):}
A task family requires a shared reward form that aligns with operator intent while adapting to evolving specifications. For instance, cooling efficiency balances energy with operational safety, applicable across scenarios with different thermal loads and SLA bounds $\psi$. Power capping prioritizes capacity constraints, adapting to both low-density ($\mu_\mathrm{min}$) and high-density ($\mu_\mathrm{max}$) server rooms with varying SLA limits ($\psi_\mathrm{min}, \psi_\mathrm{max}$). 
The challenge is: \textit{how to generate reward forms $\mathcal{R}_{\psi}$ such that the same structure generalizes across all specifications $(\mu, \psi) \sim p(\mu) \times p(\psi)$?} 

\textbf{Specification-to-Policy Latency (C2): }
Traditional DRL trains separate agents per specification $\mathcal{M}_i$, requiring hour-level retraining. 
Specification changes lead to DRL agent performance degradation. 
In particular, installing additional servers increases heat load and modifies thermal dynamics, creating a dynamics transfer where the actual state transition probability $\mathcal{T}'$ differs from the nominal model $\mathcal{T}$ used during training. 
Besides, SLA requirements evolve as tenant needs change, yielding a new reward $\mathcal{R}'$ with modified SLA setpoints different from $\mathcal{R}$. 

\section{DCoPilot Design}
\label{sec:design}
In this section, we present the system design of DCoPilot, which integrates LLM-based symbolic and hypernetwork-based parametric generation of DC control policies.

\subsection{Specification Encoding \& Scene Synthesis}
A DC operational specification $(\mu, \psi)$ defines an operational scenario, which is set according to realistic DC operating ranges. DCoPilot supports interpolation within the trained specification envelope.
The scene configuration $G$ describes a DC SimReady scene with each asset paired with a calibrated SimReady model. 
The dynamics parameter $\mu$ captures thermal load variations (e.g., server density, power consumption) that shift transition dynamics $\mathcal{T}_{\mu}$. 
The SLA parameter $\psi$ encodes tenant requirements (e.g., server inlet temperature, room humidity) that shape a reward function $\mathcal{R}_{\psi}$. Each task family corresponds to an operator objective $I_{\text{ops}}$ expressed in natural language (e.g., "minimize cooling power consumption while maintaining the air temperature of data hall A"). 

DCoPilot decomposes specification-to-policy generation into: (i) LLM translates $(G, I_{\text{ops}}) \rightarrow R_{\psi}(\cdot)$, and (ii) hypernetwork maps $(\mu, \psi) \rightarrow \theta$.
To enable generalization, DCoPilot synthesizes diverse training scenarios by sampling specification distributions. 
Starting from the current calibrated scene $G_0$, we generate scene variants by modifying asset configurations. Each scene $G_i$ corresponds to a unique dynamics parameter $\mu_i$ (e.g., $G_1$ with $\SI{80}{\kilo\watt}$ load has $\mu_1$, $G_2$ with $\SI{120}{\kilo\watt}$ has $\mu_2$), yielding $\bm{\mu} = \{\mu_1, \ldots, \mu_m\}$. 
For SLAs, we sample $\bm{\psi} = \{\psi_1, \ldots, \psi_n\}$ with varying bounds (e.g., $T_{\max} \in \{$\SI{22}{\celsius}$, $\SI{25}{\celsius}$, $\SI{27}{\celsius}$\}$). 
The Cartesian product $\bm{\mu} \otimes \bm{\psi}$ yields a set of task embeddings $\mathbf{e}$ to define MDP instances as:
\begin{equation}
    \mathbf{e} = \{(\mu_i, \psi_j) \mid i=1,\ldots,m, \; j=1,\ldots,n\},
\end{equation}
where each $(\mu_i, \psi_j)$ pairs scene $G_i$ with SLA $\psi_j$.
By training on this finite set, the hypernetwork learns to interpolate across the continuous parameter space, thereby generating policies for specifications within the sampled range.

\subsection{LLM-Based Symbolic Reward Generation}
\subsubsection{Family reward design}
To address \textbf{C1}, DCoPilot employs an LLM to synthesize a shared reward form that generalizes across the MDP family $\mathscr{M}$. Figure~\ref{fig:workflow} shows the symbolic reward generation workflow. Unlike piecewise reward design for individual MDPs $\mathcal{M}_i$, family reward synthesis produces a parameterized structure applicable across specification distributions $(\mu, \psi) \sim p(\mu) \times p(\psi)$.

Given scene configuration $G(\mu)$ with environment specification $\mu$ and operator objective $I_{\text{ops}}$ in natural language, the Reward LLM generates executable reward code. We contextualize $I_{\text{ops}}$, scene $G$, and DRL training configuration $I_{\text{cfg}}$ using prompt template $C_{\text{reward}}$ as $Q_{\text{reward}} = C_{\text{reward}}(I_{\text{ops}}, I_{\text{cfg}}, G)$. The Reward LLM produces $N$ candidate reward forms (\ding{202}):
\begin{equation}
    \{\bar{R}_i\}_{i=1}^N \sim \Theta_\mathrm{reward}(Q_\mathrm{reward}),
\end{equation}
where $\bar{R}_i$ represents the discrete form of a reward function. An example of reward form is shown in Appendix~\ref{apx:reward}.
Each reward candidate combines a list of parameterized SLA setpoints $\bm{\psi} = \{\psi_1,\psi_2,\dots,\psi_m\}$.
Thus, a family of reward functions can be denoted as:
\begin{equation}
    \mathscr{R}=\{R_{i}|R_{i}(\mathbf{s}|\bm{\psi})\},
\end{equation}
where the reward function $R_i$ shares the same functional form $\bar{R}$ with others but with a different parameter $\psi$.
The goal is to identify an effective reward form $\bar{R}^*$ that maximizes policy performance across the specification distribution, which will serve as the foundation for subsequent policy training and hypernetwork distillation.

\subsubsection{Evolutionary optimization}
To enable iterative reward refinement, we evaluate candidates on boundary conditions that stress-test generalization across the specification distribution. 
Rather than testing on random samples, we choose the minimum and maximum values of both environment and SLA parameters, denoted as $\langle \psi_{\min}, \psi_{\max} \rangle$ and $\langle\mu_{\min}, \mu_{\max}\rangle$. The Cartesian product yields boundary task embeddings as:
\begin{equation}
    \bm{\textbf{e}}_{\text{bound}} = \langle\psi_{\min}, \psi_{\max}\rangle \otimes \langle\mu_{\min}, \mu_{\max}\rangle.
\end{equation}
For each candidate $\bar{R}_i$ and boundary condition $(\mu, \psi) \in \mathbf{e}_{\text{bound}}$, we train a DRL policy by maximizing expected return (\ding{203}):
\begin{equation}
\label{eq:bnd-train}
\pi^{*}_{i,\mu,\psi} \in \arg\max_{\pi} \mathbb{E}_{\tau \sim (\pi; G(\mu))}\left[\sum_{t=0}^{T}\gamma^{t} R_{i}(s_t,a_t|\psi)\right].
\end{equation}
Rolling out each trained policy produces trajectories (\ding{204}):
\begin{equation}
\label{eq:bnd-rollout}
\pi^{*}_{i,\mu,\psi}(a|s) \xrightarrow{\text{Rollout on } G(\mu)} \hat{\tau}_{i}(\mu,\psi), \quad \forall(\mu,\psi) \in \mathbf{e}_{\text{bound}}.
\end{equation}
From each trajectory $\hat{\tau}_i(\mu, \psi)$, DCoPilot computes SLA violation costs and objective scores. These metrics are aggregated into the context describing performance at boundary conditions (\ding{205}). 
The top-$k$ reward forms and their performance summaries are appended to the prompt context, enabling the Reward LLM to generate improved candidates in the next iteration. This evolutionary process continues until convergence, selecting $\bar{R}^*$ as the final reward form for policy pool curation.

\subsubsection{Trajectory pool curation}
With the best reward form $\bar{R}^*$ selected, DCoPilot curates a diverse policy pool by training DRL agents across the sampled specification distribution. 
The set of task vectors for tasks across different environments and rewards is $\textbf{e} = \bm{\mu} \otimes \bm{\psi}$.
For each $e=(\mu,\psi)\in\textbf{e}$, we train a policy with reward $R(\psi)$ under the SimReady environment $G(\mu)$ (\ding{206}):
\begin{equation}
\label{eq:task-train}
\pi^{*}_{e} \in \arg\max_{\pi} \mathbb{E}_{\tau \sim (\pi; G(\mu))}\left[\sum_{t=0}^{T}\gamma^{t} R^*(s_t,a_t|\psi)\right],
\end{equation}
where $R^* = \bar{R}^*(\cdot|\psi)$ instantiates the shared reward form with SLA parameters $\psi$, and $G(\mu)$ denotes the scene corresponding to dynamics $\mu$. 
The key assumption underlying our approach is that converged DRL training establishes a mapping from each MDP $\mathcal{M}_e \in \mathscr{M}$ to a near-optimal policy $\pi_e^*: \mathcal{S} \rightarrow \text{Dist}(\mathcal{A})$. Since we parameterize transition dynamics $\mathcal{T}_\mu$ and reward function $R_\psi$ with $\mu$ and $\psi$, each near-optimal policy can be expressed as $\pi_e^{*}(a|s, \mu_e, \psi_e)$ conditioned on the task embedding $e=(\mu_e, \psi_e)$ for MDP $\mathcal{M}(\mu_e, \psi_e)$. This parameterization enables the hypernetwork to learn the implicit mapping from task embeddings to policy structures.
We rollout each trained policy to collect demonstration trajectories (\ding{207}):
\begin{equation}
    \pi_e^{*}(a|s) \xrightarrow{\text{Rollout on } G(\mu)} \hat{\tau}_e = \{(s_t, a_t)\}_{t=0}^{T}.
\end{equation}
Aggregating across all embeddings yields the trajectory pool:
\begin{equation}
    \mathbf{T} = \{\hat{\tau}_e \mid \hat{\tau}_e , ~\forall e \in \mathbf{e}\}.
\end{equation}
Each trajectory $\hat{\tau}_e$ paired with embedding $e = (\mu, \psi)$ provides supervised examples for hypernetwork distillation, encoding the specification-to-behavior mapping across the MDP family.

\subsection{Hypernetwork-Based Parametric Network Generation}
\label{sec:policy_distill}

\subsubsection{Meta policy architecture}
\begin{figure}[t]
    \centering
    \includegraphics[width=\linewidth]{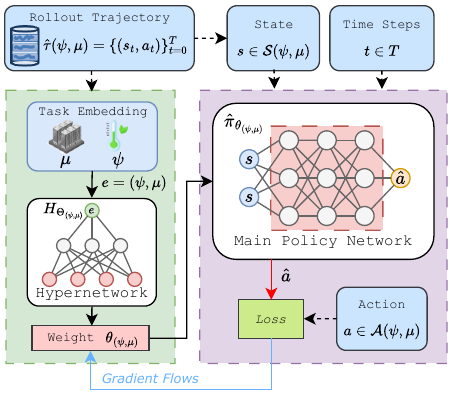}
    \vspace{-2em}
    \caption{Overview of the hypernetwork-based policy distillation framework. DCoPilot uses the near-optimal trajectory pool to train the hypernetwork.}
    \label{fig:hypernet}
\end{figure}

To address \textbf{C2}, DCoPilot employs a hypernetwork that generates policy weights conditioned on specification embeddings. 
As illustrated in Figure~\ref{fig:hypernet}, the architecture consists of a hypernetwork $H_\Theta$ and a main policy network $\pi_\theta$.
The hypernetwork $H_\Theta$ is a multi-layer perceptron that takes task embedding $e=(\mu,\psi)$ as input and outputs the full weight set $\theta$ for the main policy. The embedding encoder first projects $\mu$ and $\psi$ into fixed-dimensional representations through separate embedding layers, then concatenates them to form the joint embedding $e \in \mathbb{R}^d$. 
The hypernetwork processes this embedding through several fully-connected layers, producing weight tensors for each layer of the main policy network.

The main policy network $\pi_\theta$ follows a standard actor architecture for continuous control, where it maps observations $s \in \mathcal{S}$ to action distributions over $\mathcal{A}$.  
The main policy architecture remains fixed, where only its parameters are generated by $H_\Theta$ conditioned on task embeddings:
\begin{equation}
\theta = H_\Theta(e), \quad a \sim \pi_{\theta}(a|s).
\end{equation}
This design enables parameter sharing across the MDP family. Rather than training independent policies for $\mathcal{M}_i \in \mathscr{M}$, DCoPilot learns a single hypernetwork capturing shared control structures for the MDP family $\mathscr{M}$. The hypernetwork implicitly encodes how specifications $(\mu, \psi)$ should modulate policy behavior, facilitating zero-shot generalization to new task embeddings.

\subsubsection{Distillation learning}
Let $\mathbf{T}$ denote the curated pool of near-optimal trajectories, where each trajectory $\tau_i^*=\{(s_{i,t},a_{i,t})\}_{t=0}^{T}$ was obtained under the task $e_i=(\mu_i,\psi_i)$. 
We cast hypernetwork training as a supervised imitation problem, where for each trajectory $\tau_i^*$ the hypernetwork should produce a policy whose action distribution matches the demonstrated actions along the observed states. 
The hypernetwork training is:
\begin{equation}
\mathcal{L}(\Theta) = -\sum_{i,t}\log \pi_{H_{\Theta}(e_i)}(a_{i,t}\mid s_{i,t}),
\end{equation}
where $e_i \in \mathbf{e}$ encodes the environment and reward parameters $(\mu_i, \psi_i)$ corresponding to trajectory $\tau^*_i$.
DCoPilot uses the hypernetwork to generalize across different DC environments and reward parameters, enabling zero-shot adaptation of policies for unseen operational specifications.

\subsubsection{Zero-shot deployment}
After distillation, DCoPilot generates the meta policy hypernetwork $H(\Theta)$. 
For online adaptation to new scenes and SLAs, we first identify the scene feature $\mu'$ from monitoring IoT data and set the SLA setpoints $\psi'$.
The hypernetwork $H(\Theta)$ uses new embedding $e'=(\mu',\psi')$  to generate the main policy weight $\theta'$ as:
\begin{equation}
\theta' = H(\Theta,e').
\end{equation}

The generated policy $\pi_{\theta'}$ deploys immediately for online control, eliminating specification-to-policy latency. This zero-shot adaptation does not require environmental interaction, gradient updates, or retraining, addressing the minute-scale control requirements discussed in Section~\ref{sec:pre}. The hypernetwork leverages learned structure across the MDP family to generalize to new specifications through interpolation within the trained embedding space.

\section{Experiments}
\label{sec:eval}
We evaluate DCoPilot on multiple DC cooling control task families, comparing its policies with baselines and ablations to demonstrate the effectiveness of LLM-based symbolic and hypernetwork-based parametric generation.

\begin{table*}[t]
\centering
\caption{Average violation cost and objective score of the actions given by baselines and our method. We use column \textbf{Action} as the direct model type to give the actions, \textbf{Transfer} as the considered dimension of transfer learning in the process. The dual value means the violation cost with the objective score. A lower cost and score are preferable across all tasks. The best method with the lowest cost and score is highlighted in \textbf{bold} text.  }
\vspace{-0.5em}
\label{tab:rew_evolution}
\begin{tabular}{@{}c|cc|c|cc|cc@{}}
\toprule
\multirow{2}{*}{\textbf{Method}}         & \multicolumn{2}{c|}{\textbf{Control}} & \textbf{Zone 1} & \multicolumn{2}{c|}{\textbf{Zone 2}} & \multicolumn{2}{c}{\textbf{Indoor + Outdoor Plant}} \\ \cmidrule(l){2-8} 
                                         & \textbf{Action}  & \textbf{Transfer}  & (a) (s1/o1)$\downarrow$     & (b) (s1/o1) $\downarrow$           & (c) (s1/s2/o1) $\downarrow$          & (d) (s1/o2) $\downarrow$   & (e) (s1/o1/o2) $\downarrow$         \\ \midrule
\multicolumn{1}{c|}{(i) LLM4PID}             & PID              & N/A                & 0.05/1.75            & 0.36/1.69                      &         0.36/0.95/1.69           & 0.34/3.94          & 0.34/1.64/3.94           \\
\multicolumn{1}{c|}{(ii) LLM-as-Policy}       & LLM              & N/A                & 5.90/1.60            & 3.46/1.67                      & 3.32/3.22/1.66                     & 3.13/3.83              & 3.27/1.58/3.80                    \\
\multicolumn{1}{c|}{(iii) EvoReward}              & DRL              & No                 & 1.18/1.72            & 3.09/1.64                      & 0.99/1.69/1.72                     & 8.78/3.96          & 4.05/1.76/4.21           \\ 
\midrule
\multicolumn{1}{c|}{ABL (w/o bound)}    & Meta             & Env+Rew            & 1.42/1.73                & 0.82/1.72                      & 0.44/0.80/1.76                     & 0.07/3.86              & 0.32/1.58/3.77                    \\
\multicolumn{1}{c|}{\textbf{DCoPilot}}                & Meta             & Env+Rew            & \textbf{0.01/1.63}   & \textbf{0.02/1.68}                      & \textbf{0.01/0.00/1.69}                     & \textbf{0.18/3.80}              & \textbf{0.19/1.59/3.80}  \\ \bottomrule
\end{tabular}
\end{table*}

\subsection{Experimental Setup}
\subsubsection{Task design}
We consider two SLAs in today's DC operations:
(s1) \textit{temperature}, which maintains the server room temperature within specified bounds, and
(s2) \textit{humidity}, which maintains air humidity in some tropical DCs for the contained ITE. 
Notably, the SLAs only affect indoor facility operations as the serving ITE is contained within the server room.
Then, we consider two optimization objectives in DC operations:
(o1) \textit{power consumption} of HVAC, which is a major source of DC's energy consumption, and
(o2) \textit{water consumption} of chillers, which is also one of the core resource-saving issues.
We combine two SLA targets and two objectives, yielding five task families denoted by (a)-(e).
Specifically, we consider the DC as an evolving environment, where the number of installed server units is changing. 
Task family (a) is applied with the SLA (s1) and the objective (o1) for a single-CRAC server room named \textit{Zone 1}. 
For task family (a), the observation space captures indoor zone-level states, including the IT workload schedule, zone air temperature, and supply air temperature. The action space consists of continuous control inputs to the CRAC unit, including supply air temperature setpoints and fan airflow rates.
Task family (b)-(c) are applied with both (s1) and (s2) and the indoor objective (o1) for a server room named \textit{Zone 2}, including dehumidifiers. 
For task families (b) and (c), the observation space augments that of task family (a) with relative humidity measurements.
Task family (d)-(e) are applied with the SLA (s1) and both (o1) and (o2) for the indoor Zone 1 with outdoor plant facilities denoted as \textit{Indoor+Outdoor Plant}.
For task families (d) and (e), the action space is extended to include outdoor plant-level states related to the chilled water system and cooling tower operations.

\subsubsection{Metrics}
For each task family, we use five dynamics parameters $\bm{\mu}$ by server installations and five SLA parameters $\bm{\psi}$ by adjusting SLA bounds to synthesize the operational specifications. 
We set the violation thresholds to 
$\varepsilon_{\mathrm{s1}}=1\,^{\circ}\mathrm{C}$ for the temperature SLA (s1) and 
$\varepsilon_{\mathrm{s2}}=5\,\%\mathrm{RH}$ for the humidity SLA (s2), and compute the violation cost accordingly.
We report the time-averaged power usage effectiveness (PUE) for (o1) and water usage effectiveness (WUE) for (o2).
The single objective score shown in each cell is the mean over the evaluation horizon.
Each control decision in our experiments represents a 15-minute interval.

\subsubsection{Baselines \& ablation settings}
We consider four of the existing baseline methods for LLM-generated control policies and one ablation setting.
We evaluate baselines and our method on the above task families to exhibit generative policy performance and online adaptation cost.
The baselines include 
(i) \textit{LLM4PID} \cite{koziolek2024llm} uses RAG-based LLM to write PID control codes for certain targets.
(ii) \textit{LLM-as-Policy} \cite{song2023pre} uses historical data as in-context examples with few-shot prompting for direct industrial control.
(iii) \textit{EvoReward} \cite{ma2023eureka} exploits the LLMs to perform evolutionary optimization over reward code and to train the corresponding policy via reinforcement learning.
We also include one ablation setting to show the effects of family reward code generation, which is \textit{ABL (w/o bound)} removes the boundary trajectory information during the reward evolution.
We conduct the experiments with \texttt{GPT-3.5-Turbo} and fix the LLM sampling parameters (temperature=0.7, $N$ = 5).

\begin{figure}
    \centering
    \includegraphics[width=\linewidth]{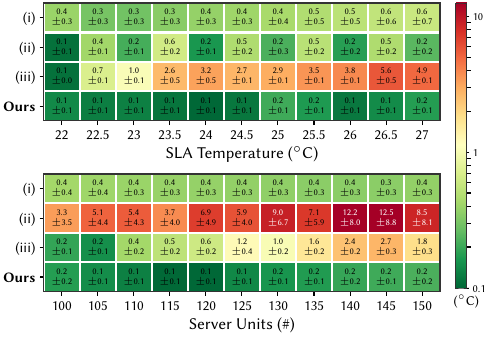}
    \vspace{-2em}
    \caption{The temperature SLA error of the policies generated by LLM-based generative policies (i)-(iii) and DCoPilot. }
    \label{fig:infer_sla_env_temp}
\end{figure}

\subsection{Generative Policy Evaluation}
\subsubsection{Performance across generative approaches}
We evaluate how different generative methods handle within-family specification variations by training on one representative task and testing across all specifications in each family. Table~\ref{tab:rew_evolution} presents the average violation costs and objective scores, where lower values indicate better performance.
DCoPilot achieves the best performance across all task families, maintaining violation costs below \SI{0.2}{\celsius} while baseline methods range from \SI{0.77}{\celsius} to \SI{8.78}{\celsius}. This significant improvement demonstrates the power of combining two forms of generative AI, i.e., LLMs for semantic understanding and hypernetworks for parametric policy generation. The LLM translates natural language specifications into appropriate reward structures, while the hypernetwork generates policy weights.
Direct LLM generation methods reveal fundamental limitations when applied to control tasks. LLM-as-Policy produces violation costs from \SI{3.13}{\celsius} to \SI{5.90}{\celsius} because token-by-token generation cannot maintain control consistency across time steps. Each action is generated independently, causing unstable control behaviors. LLM4PID performs better on constraints but poorly on objectives, achieving scores of 1.64/3.94 versus DCoPilot's 1.59/3.80. While LLMs can tune PID parameters, they cannot overcome the inherent limitations of classical control for multi-objective optimization.
Task-specific DRL methods cannot generalize across specifications. EvoReward shows catastrophic performance on complex tasks, reaching \SI{8.78}{\celsius} violations in task family (d), because each agent learns fixed behaviors tied to its training configuration. 
These methods cannot adapt to new specifications without complete retraining.
Our ablation study confirms the importance of guided reward generation. Removing boundary trajectory guidance increases violations by 5 to 70 times across different families. The boundary trajectories help the LLM understand safety boundaries when generating rewards, transforming abstract SLA requirements into concrete optimization objectives that the hypernetwork can learn effectively. 

\begin{figure}[t]
    \centering
    \includegraphics[width=\linewidth]{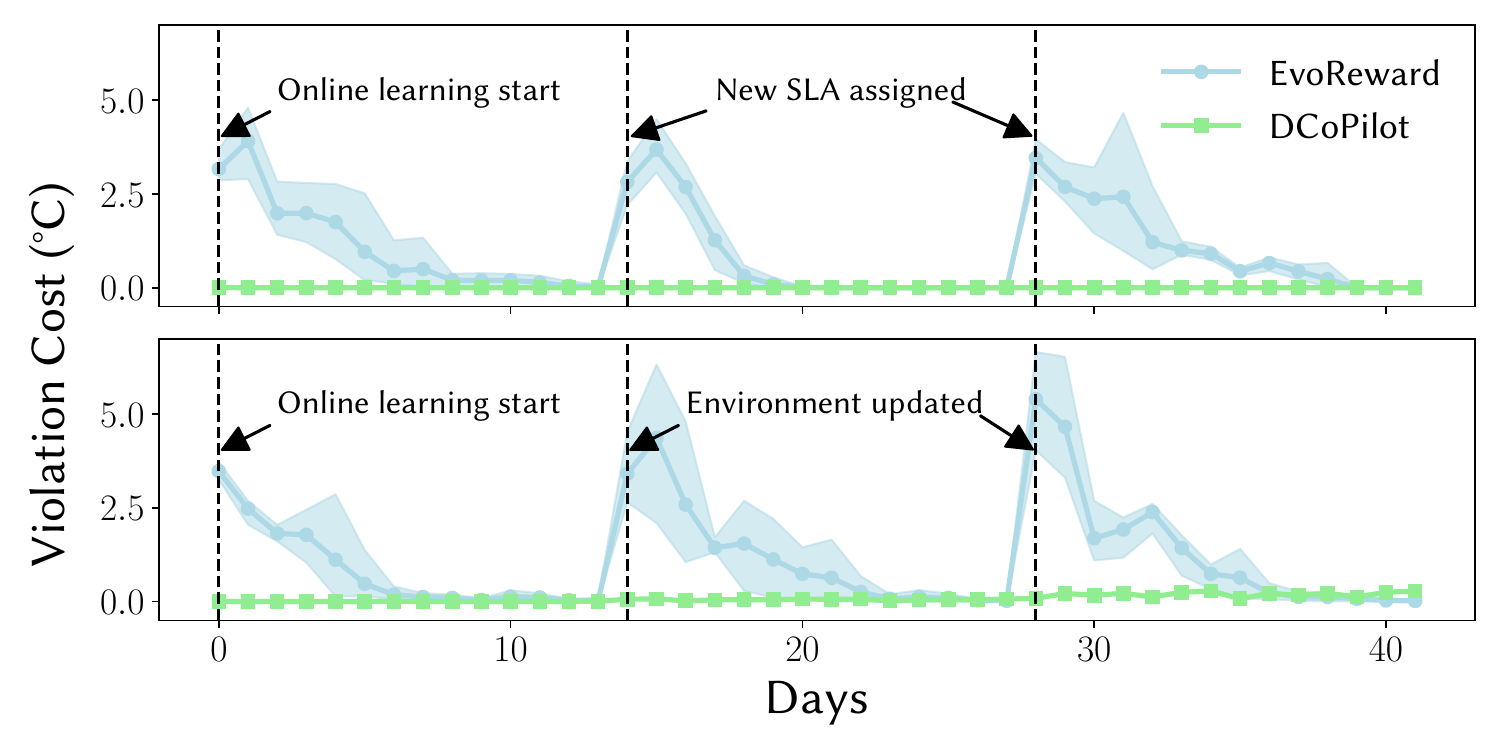}
    \vspace{-2.5em}
    \caption{\textit{Day-level} online adaptation violation cost of assigning new SLAs and updating environments in task family (a) by EvoReward and DCoPilot.}
    \label{fig:online_overhead}
\end{figure}

\subsubsection{Robust within-family generalization}
Figure~\ref{fig:infer_sla_env_temp} visualizes why DCoPilot succeeds where task-specific methods fail across two critical generalization dimensions, systematically evaluating all four approaches on task family (a) across varying SLA temperature bounds (\SI{22}{\celsius}-\SI{27}{\celsius}) and server densities (100-150 units).
LLM4PID maintains good performance in SLA generalization at \SI{0.43}{\celsius} mean absolute error (MAE) and similarly in environment generalization at \SI{0.38}{\celsius} MAE.
While generating PID codes for DC control can fulfill SLA constraints, it is incapable of objective optimization as shown in Table~\ref{tab:rew_evolution}.
LLM-as-Policy achieves \SI{0.33}{\celsius} error across SLA variations through retrieval from its demonstration database. However, it catastrophically fails in environment generalization with \SI{7.24}{\celsius} mean error and 100\% violations at 140+ server units.
This RAG-based approach cannot find relevant examples for new specifications outside its history distribution, and thus generates risky actions in unseen environments.
In contrast, DCoPilot maintains consistent sub-degree temperature errors across all tested configurations, achieving mean errors of \SI{0.43}{\celsius} for SLA generalization and \SI{0.38}{\celsius} for environment generalization with standard deviations below \SI{1}{\celsius}. This consistent performance stems from the hypernetwork architecture, which learns the mapping from specifications to appropriate policy weights rather than fixed behaviors tied to specific training configurations.
This architectural difference explains the performance gap in Table~\ref{tab:rew_evolution}.

\subsubsection{Day-level performance under online specifications}
To evaluate how different generative methods handle realistic operational changes, we simulate a 40-day DC operation scenario with specification changes occurring at days 14 and 28. These changes represent common operational events, i.e., initiating online learning for initial deployment, assigning new SLA requirements, and updating server infrastructure. Figure~\ref{fig:online_overhead} presents the accumulated violation costs during these transitions.
The results reveal critical differences in the adaptation capacities of the generative approaches. EvoReward exhibits severe violation spikes at each specification change, with costs reaching up to \SI{4}{\celsius} when new SLAs are assigned or environment updates. 
These spikes occur because EvoReward must retrain entirely new agents for each specification, leaving the system effectively uncontrolled during a 10-day online training period. During this window, the previously trained agent continues operating with obsolete parameters, causing temperature deviations that accumulate into significant violations.
In contrast, DCoPilot maintains near-zero violations throughout all specification changes, with costs remaining below \SI{0.1}{\celsius} even during transitions. This stability stems from the hypernetwork's ability to instantly generate specification-appropriate policies without any online adaptation period. When new SLAs are assigned at day 14, DCoPilot immediately produces updated policy weights that reflect the changed temperature bounds. Similarly, when server capacity changes at day 28, the hypernetwork adapts the control strategy to account for the altered thermal dynamics.

\begin{figure}[t]
    \centering
    \includegraphics[width=\linewidth]{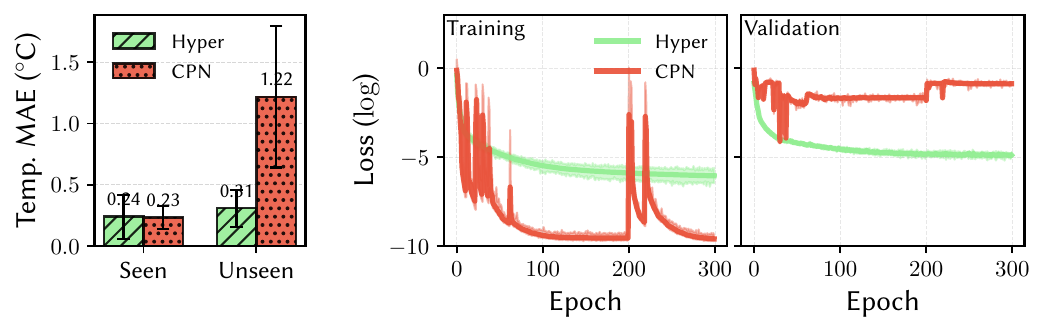}
    \vspace{-2.5em}
    \caption{(Left) Temperature MAE and (right) loss comparison between Hypernetwork and CPN on task family (a).}
    \label{fig:all_tasks_comparison}
\end{figure}

\subsection{Minute-Level Response with Hypernetwork}
\subsubsection{Policy-Parameter Generation vs. Action-Level Conditioning}
A natural alternative to hypernetwork-based policy generation is to directly condition the policy on the control specification, i.e., learning a conditional policy of the form $\pi(a \mid s, e)$ as in conditional policy networks (CPNs). 
Figure~\ref{fig:all_tasks_comparison} shows that using hypernetwork achieves a lower temperature MAE at 0.31 compared to CPN at 1.22 on interpolated but unseen specifications.
While CPN exhibits a faster decrease in training loss, it suffers from consistently higher validation loss. 
This behavior arises from its inductive bias, where CPN performs action-level interpolation in the joint state–specification space and implicitly assumes smooth transitions of the optimal action with respect to specification changes. However, in DC control, even interpolated specification changes can induce substantial shifts in the underlying state distribution, rendering such interpolation unreliable. 
In contrast, the hypernetwork explicitly models how control specifications transform policies by learning a mapping $e \mapsto \theta_e$, followed by execution via $\pi_{\theta_e}(a \mid s)$. 
By performing interpolation at the parameter level, the hypernetwork generates coherent task-specific policies.

\begin{figure}[t]
    \centering
    \includegraphics[width=\linewidth]{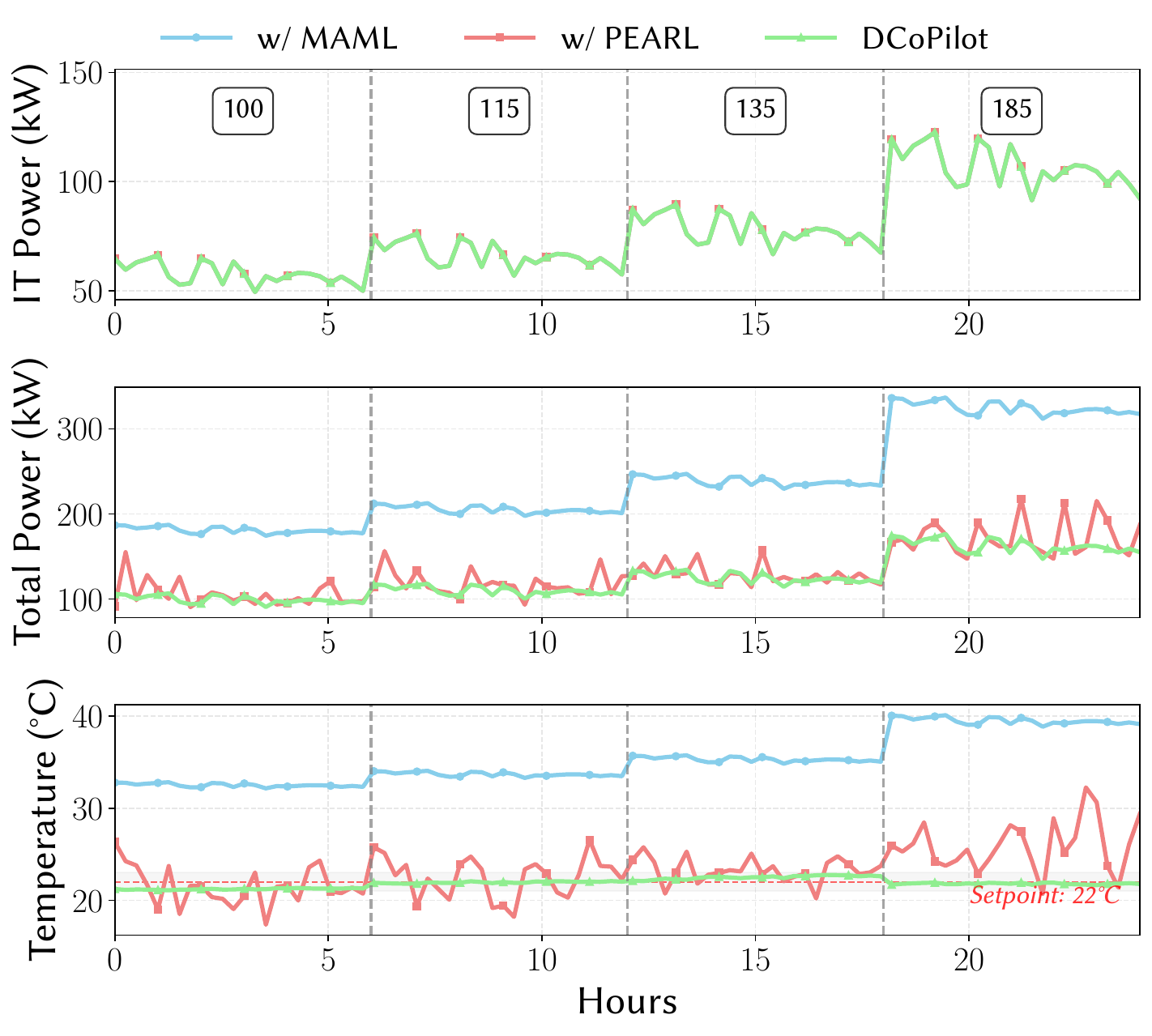}
    \vspace{-2.5em}
    \caption{\textit{Minute-level} online adaptation violation cost of new specifications (e.g., server units=100, 115, 135, 185) in task family (a) by MAML, PEARL, and hypernetwork for parametric policy weight generation of DCoPilot.}
    \label{fig:meta_learning}
\end{figure}

\subsubsection{Comparison with Meta-Learning Methods}
We further compare DCoPilot against two prominent meta-learning baselines, MAML and PEARL, under identical training conditions using the same LLM-generated reward form and specification distribution. 
We evaluate zero-shot transferability across four consecutive specification changes (i.e., 100, 115, 135, and 185 servers) over a 24-hour control horizon, deploying all methods without any online fine-tuning to simulate realistic operational scenarios where server workload variations or hardware reconfigurations require immediate policy adaptation.

Figure~\ref{fig:meta_learning} shows simulation results for MAML, PEARL, and the hypernetwork across three dimensions, including IT power consumption, total power consumption, and zone air temperature. Throughout four specification transitions, DCoPilot demonstrates superior control stability across all metrics. 
For total power efficiency, DCoPilot maintains consistently lower power consumption, indicating more effective cooling coordination. Most notably, in temperature regulation, DCoPilot maintains the DC temperature stably within the SLA-acceptable range around the \SI{22}{\celsius} setpoint throughout all specification transitions, whereas both MAML and PEARL exhibit larger temperature deviations, particularly during workload changes.
The superior performance comes from the hypernetwork directly generating task-specific parameters via explicit specification-to-policy mapping, unlike MAML’s gradient-based adaptation and PEARL’s probabilistic context inference. The generative control paradigm removes adaptation latency and performance issues seen in meta-learning when specifications exceed training coverage.

\subsection{Role of LLM in Meta Policy Learning}
\label{sec:exp_ablation}
\subsubsection{Family reward form enables hypernetwork convergence}
\begin{figure}[t]
    \centering
    \includegraphics[width=\linewidth]{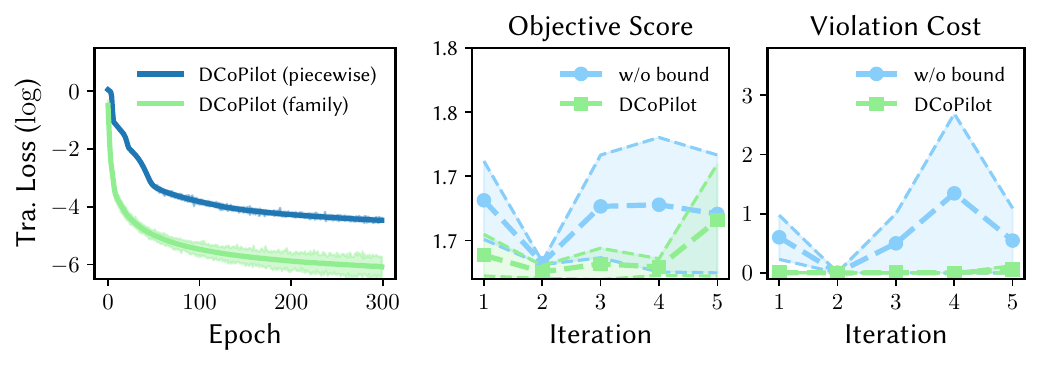}
    \vspace{-2.5em}
    \caption{(Left) Training loss convergence comparing unified reward generation against piecewise reward generation. (Middle, Right) Evolution of objective score and violation cost across reward refinement iterations. All experiments conducted on task family (a).}
    \label{fig:llm_in_meta}
\end{figure}

We evaluate the role of LLM-generated family reward forms for effective hypernetwork training by comparing DCoPilot's unified reward generation against piecewise reward generation, where separate reward functions are created for each specification. 
In the piecewise reward generation approach, the LLM generates separate reward functions for each individual specification, trains corresponding DRL agents, and then distills these agents into a hypernetwork. 
Figure~\ref{fig:llm_in_meta} presents the training loss on task family (a). 
DCoPilot with unified reward generation achieves stable convergence in hypernetwork training, demonstrating effective interpolation between different specifications. The piecewise reward generation baseline fails to converge, exhibiting 100x higher loss throughout training.
The LLM generates unified reward forms covering all specifications, which ensures all trained policies share the same MDP family.

\subsubsection{Boundary trajectories stabilize reward evolution}
We evaluate the effectiveness of boundary trajectory guidance during LLM-based reward evolution by comparing DCoPilot against the ablation baseline without boundary information. Reward quality at each iteration is measured by the average violation cost and objective score of the best candidate, evaluated across five different LLM sampling seeds. 
Figure~\ref{fig:llm_in_meta} presents the evolution trends on task family (a).
DCoPilot demonstrates rapid convergence to near-zero violation costs within the first few iterations and maintains stable performance throughout subsequent rounds. In contrast, the baseline without trajectories of boundary specifications exhibits fluctuating violation costs, occasionally producing reward candidates that cannot adequately encode safety constraints. Both methods achieve comparable objective scores, but DCoPilot's consistent constraint satisfaction demonstrates more reliable reward generation. This stability stems from the boundary trajectory evaluation strategy, where testing reward candidates on extreme specifications provides comprehensive stress-testing. This boundary-aware feedback guides the Reward LLM to generate symbolic reward forms that generalize across the full specification distribution.

\section{Discussion}
\label{sec:discuss}
DCoPilot’s dual-generative architecture addresses a core challenge in DC operations, where specification changes occur faster than conventional policy updates.
Although it shows strong zero-shot generalization across diverse control tasks, four key challenges remain.
First, bridging the gap between simulated and real DC dynamics is difficult due to mismatches in airflow, heat transfer, and actuator latency~\cite{cao2025transforming}.
Hypernetworks trained on synthetic data may falter with sensor noise or unmodeled system couplings in real settings. Accurate SimReady digital twins need continuous recalibration with evolving hardware and workloads~\cite{vering2021digital}. 
Future work should focus on domain adaptation methods to ensure robust policy transfer to live environments.
Second, neither the LLM-based reward generator nor the hypernetwork offers formal convergence guarantees under non-convex optimization.
LLM outputs are generated stochastically rather than optimized by gradient descent, causing variability in reward definitions~\cite{novikov2025alphaevolve}.
The hypernetwork, trained over a distribution of tasks instead of a single objective, may exhibit unstable convergence~\cite{sarafian2021recomposing}.
Future work will explore theoretical frameworks for the convergence.
Third, this paper only considers homogeneous specification changes that can be parameterized, but real-world DC evolution often involves heterogeneous changes to the facility topology~\cite{poutievski2022jupiter}.
Addressing this limitation needs a hierarchical framework, where an LLM provides expert-guided recommendations for complex heterogeneous changes, while low-level controllers handle the detailed setpoints.
Future work will also explore heterogeneous topological changes via modular hypernetworks, where each generates policy parameters for a specific subsystem. 
Moreover, our approach only provides soft constraint adherence through reward shaping at training time rather than worst-case guarantees at serving time. 
Incorporating explicit serving-time safety layers~\cite{wang2024green} is an important direction for future work.

\section{Conclusion}
\label{sec:conclude}
This paper presents DCoPilot, a hybrid framework that synergizes LLM-based symbolic reward generation with hypernetwork-based parametric policy generation to address specification-to-policy latency in dynamic DC control. 
This work establishes a new paradigm for generative CPS control, where semantic understanding from pre-trained language models combines with parametric precision from neural policy generators to enable automated and adaptive operations in mission-critical, evolving industrial environments.

\section*{Acknowledgment}
This research is supported in part by the RIE2025 Industry Alignment Fund – Industry Collaboration Projects (IAF-ICP) (Award I2301E0026), administered by A*STAR, in part by Alibaba Group and NTU Singapore through Alibaba-NTU Global e-Sustainability CorpLab (ANGEL), in part by the National Research Foundation, Singapore, under its AI Singapore Programme (AISG Award No: AISG4-GC-2023-006-1B), and in part by Singapore Ministry of Education under its AcRF Tier-1 grant RT14/22.

\balance
\bibliographystyle{IEEEtran}
\bibliography{sample-base}

\clearpage

\appendix

\section{Offline Compute Cost and Scalability Analysis}

This appendix provides a quantitative analysis of the offline computational cost of the proposed specification-driven policy generation pipeline. While the main paper emphasizes zero-shot online adaptation with no additional training at deployment time, the offline cost is incurred once per task family and is amortized across all downstream deployments.
The offline pipeline includes (i) LLM-assisted reward initialization and verification, (ii) training and rollout of expert DRL policies across sampled control specifications, and (iii) hypernetwork distillation from the collected trajectories. 

\subsection{Compute Cost Breakdown}

Table~\ref{tab:spec_resolution_cost} reports a detailed breakdown of the offline compute time for task family (a) under 8 and 32 control specifications. All experiments are conducted on a single experimental server.
For the 8-specification setting, the total makespan is 2244.49 seconds. The offline pipeline can be divided into two phases. 
The first phase is LLM-assisted reward evolution, which includes reward initialization (\ding{202}), verification (\ding{203}-\ding{204}), evaluation (\ding{205}), and sampling of new reward candidates (\ding{202}). 
The second phase is DRL-based policy learning, comprising policy training (\ding{206}), policy rollout (\ding{207}), and hypernetwork training. 
The reward evolution phase accounts for 56.5\% of the total makespan (1267.63 seconds), while the policy learning phase accounts for the remaining 43.5\% (976.85 seconds). Within each phase, the dominant costs are reward verification (863.76 seconds) and policy training (479.11 seconds), respectively. Notably, hypernetwork training accounts for only 4.1\% of the total makespan (91.41 seconds).

\begin{table}[h]
\centering
\caption{Offline compute makespan under 8 and 32 specification resolutions.}
\label{tab:spec_resolution_cost}
\begin{tabular}{lr|r}
\toprule
 & \textbf{8 Specs} & \textbf{32 Specs} \\
\cmidrule(lr){2-3}
\textbf{Stage} 
 & \textbf{Time (s)} 
 & \textbf{Time (s)} \\
\midrule
Initialize rewards (\ding{202})         
 & 235.20  
 & 233.27 \\
Verify rewards (\ding{203}-\ding{204})   
 & 863.76  
 & 971.20 \\
Evaluation (\ding{205})                 
 & 156.25  
 & 172.69 \\
Sample new rewards (\ding{202})          
 & 12.42  
 & 13.18 \\
\midrule
Policy training (\ding{206})             
 & 479.11  
 & 445.44 \\
Policy rollout (\ding{207})              
 & 406.33  
 & 3868.33 \\
Hypernetwork training                    
 & 91.41  
 & 60.10 \\
\midrule
\textbf{Total makespan}                  
 & \textbf{2244.49}  
 & \textbf{5764.21} \\
\bottomrule
\end{tabular}
\end{table}

\subsection{Scalability with Specification Resolution}

To evaluate scalability, we increase the number of control specifications from 8 to 32. As shown in Table~\ref{tab:spec_resolution_cost}, the total offline time increases from 2244.49 seconds to 5764.21 seconds.
The cost increase is primarily driven by policy rollout, which grows from 406.33 seconds to 3868.33 seconds. The roughly 9.5$\times$ increase results from executing specifications sequentially in our current implementation. In contrast, policy training exhibits minimal growth because training across specifications is partially parallelized using multiple GPUs.
The LLM-assisted reward evolution phase exhibits nearly identical cost between the two settings. This is because the reward generation process focuses only on the extreme cases of the specification range, regardless of the intermediate resolution used for policy training.
Hypernetwork training time decreases from 91.41 seconds to 60.10 seconds at higher resolution due to faster convergence on denser, more informative trajectory datasets collected from more expert policies.

\subsection{Experimental Hardware Configuration}

Table~\ref{tab:hardware_config} summarizes the hardware configuration of our experimental server. The current experiments utilize 4 NVIDIA Tesla V100 GPUs for partial parallelization of policy training across specifications.
The current bottleneck is an artifact of our sequential implementation rather than a fundamental limitation of the approach. Since rollouts across different specifications are independent, they can be distributed across multiple workers. Similarly, policy training across specifications is inherently parallel, as each specification yields an independent optimization problem.

\begin{table}[h]
\centering
\caption{Hardware configuration of the experimental server.}
\label{tab:hardware_config}
\begin{tabular}{ll}
\toprule
\textbf{Component} & \textbf{Specification} \\
\midrule
Processor & Intel Xeon E5-2698 v4 @ 2.20\,GHz \\
CPU cores & 20 cores / 40 threads \\
CPU frequency range & 1.2--3.6\,GHz \\
Architecture & x86\_64 \\
\midrule
Memory & 251\,GB DDR4 \\
Available memory & 240\,GB \\
\midrule
GPU & 4 $\times$ NVIDIA Tesla V100-DGXS-32GB \\
GPU memory & 32\,GB (per GPU) \\
\bottomrule
\end{tabular}
\end{table}

\subsection{LLM Token Usage Analysis}

Table~\ref{tab:llm_usage} summarizes the LLM usage statistics for the reward evolution stage under the 8-specification setting. The reward evolution process runs for 5 iterations with a sampling size of 5 candidates per iteration. 
The LLM-based reward evolution uses \texttt{GPT-3.5-Turbo} with a total of 10 API requests, consuming 15,422 tokens, comprising 12,005 prompt tokens and 3,417 completion tokens. The average prompt length is approximately 4,604 characters. 
Since LLM queries occur only in the offline phase, the measured token consumption is moderate and acceptable, without affecting online latency or incurring additional inference API costs.

\begin{table}[h]
\centering
\caption{LLM usage statistics for offline reward evolution with 8 specifications for DCoPilot's training.}
\label{tab:llm_usage}
\begin{tabular}{ll}
\toprule
\textbf{Item} & \textbf{Value} \\
\midrule
LLM model & \texttt{GPT-3.5-Turbo}\\
Total LLM requests & 10 \\
Total tokens & 15,422 \\
\quad Prompt tokens & 12,005 \\
\quad Completion tokens & 3,417 \\
Average prompt length & 4,604 characters \\
\bottomrule
\end{tabular}
\end{table}

\newpage 
\onecolumn

\section{Reward Form Implementation}
\label{apx:reward}
\begin{lstlisting}[caption={A Generated Reward Function for Task Family (a)}, label={lst:reward_fn}]
def reward_fn(env, target_temp, power_weight=0.5):
    current_temperature = env.inspect_current_observation(
        "zone_air_temperature", use_unnormed=True
    )

    # Compute the absolute difference between current and desired temperature
    temperature_difference = abs(current_temperature - target_temp)

    # The reward is higher when the temperature difference is lower
    temperature_reward = 1.0 - (temperature_difference / 5.0)

    # Calculate the total power consumption
    ite_power = env.inspect_current_observation("IT_power", use_unnormed=True)
    crac_power = env.inspect_current_observation("CRAC_power", use_unnormed=True)
    chiller_power = env.inspect_current_observation("Chiller_power", use_unnormed=True)
    pump_power = env.inspect_current_observation("CHWP_power", use_unnormed=True)

    total_power = ite_power + chiller_power + crac_power + pump_power

    # Power penalty term
    power_reward = -total_power * power_weight / 100000

    # Combine rewards
    total_reward = temperature_reward + power_reward

    # Clip reward to (-1, 1)
    total_reward = max(-1, min(1, total_reward))

    return total_reward
\end{lstlisting}

\end{document}